\documentclass[Afour,sageh,times]{sagej}

\usepackage{moreverb,url}

\usepackage[colorlinks,bookmarksopen,bookmarksnumbered,citecolor=red,urlcolor=red]{hyperref}
\usepackage{graphicx}
\usepackage{multirow}
\usepackage{makecell}
\usepackage{enumitem}

\newcommand\BibTeX{{\rmfamily B\kern-.05em \textsc{i\kern-.025em b}\kern-.08em
T\kern-.1667em\lower.7ex\hbox{E}\kern-.125emX}}

\def\volumeyear{2023}

\usepackage{dirtree}
\usepackage{bm}
\usepackage{natbib}

\begin{document}
\runninghead{Le et al.}

\title{RflyMAD: A Dataset for Multicopter Fault Detection and Health Assessment\\
	%A demonstration of the \LaTeXe\ class file for \itshape{SAGE Publications}
}

\author{Xiangli Le\affilnum{1}, Bo Jin\affilnum{1}, Gen Cui\affilnum{1}, Xunhua Dai\affilnum{2} and Quan Quan\affilnum{1}}

\affiliation{\affilnum{1}Beihang University, P.R. China\\
\affilnum{2}Central South University, P.R. China}

\corrauth{Quan Quan, Beihang University
Reliable Flight Control Group,
No.37 XueYuan Road,
HaiDian District, Beijing,
100191, P.R. China.}

\email{qq\_buaa@buaa.edu.cn}

\begin{abstract}
This paper presents an open-source dataset RflyMAD, a Multicopter Abnomal Dataset developed by Reliable Flight Control (Rfly) Group aiming to promote the development of research fields like fault detection and isolation (FDI) or health assessment (HA). The entire 114 GB dataset includes 11 types of faults under 6 flight statuses which are adapted from ADS-33 file to cover more occasions in which the multicopters have different mobility levels when faults occur. In the total 5629 flight cases, the fault time is up to 3283 minutes, and there are 2566 cases for software-in-the-loop (SIL) simulation, 2566 cases for hardware-in-the-loop (HIL) simulation and 497 cases for real flight. As it contains simulation data based on RflySim and real flight data, it is possible to improve the quantity while increasing the data quality. In each case, there are ULog, Telemetry log, Flight information and processed files for researchers to use and check. The RflyMAD dataset could be used as a benchmark for fault diagnosis methods and the support relationship between simulation data and real flight is verified through transfer learning methods. More methods as a baseline will be presented in the future, and RflyMAD will be updated with more data and types. In addition, the dataset and related toolkit can be accessed through \href{https://rfly-openha.github.io/documents/4_resources/dataset.html}{https://rfly-openha.github.io/documents/4\_resources/dataset.html}.
\end{abstract}


\maketitle

\section{Introduction}
With the rapid development of the technology and related industries of UAVs, multicopters have become more popular and are commonly applied in daily lives than ever \citep{quan2017introduction}. As a result, health and flight safety of multicopters are attracting increasing attention. Researchers have long been focusing on the UAVs' fault detection and isolation (FDI) problem and many achievements have been made. Among these results, analytical redundancy are widely used for FDI problems and can be divided into Model-Based and Knowledge-Based methods \citep{fourlas2021survey}. For both kinds of methods, especially the Knowledge-Based methods, require simulation or real flight datasets of UAVs. Compared with research fields such as computer vision \citep{deng2009imagenet}, types of datasets designed for UAVs are fewer. Moreover, there is a notable scarcity of datasets that take UAVs' failures into account \citep{keipour2021alfa}.

The most common datasets about UAVs are found related to keywords such as SLAM, localization and image recognition, such as UAVid by \cite{lyu2020uavid}. The Blackbird UAV dataset provides a high rate of IMU and image data for about ten hours, which could be used for visual-inertial navigation and SLAM \citep{antonini2020blackbird}. NTU VIRAL is a dataset used for simultaneous localization and mapping \citep{nguyen2022ntu}. NASA's Open Data Portal\footnote[1]{NASA's Open Data Portal website: \href{https://data.nasa.gov/}{https://data.nasa.gov/}} records a series of datasets related to FDI and HA problems about aircraft, such as ``C-MAPSS Aircraft Engine Simulator Data'' and ``ADAPT Dataset''. Although the categories of related datasets are sufficient, most of them contains merely simulation data and focus on component-level faults. Due to the high hardware cost in real experiments, most datasets reflecting both normal and abnormal status of UAVs contains simulation data while datasets recording system level faults are scarce. ALFA is a real UAV dataset in system level for fixed wings with abnormal status to overcome this problem, as reported in  \cite{keipour2021alfa}.

Although the ALFA provides a real dataset with the fault status of UAVs, the amount of fault cases is still lower than expectation and the dataset is only suitable for fixed wings. The Rfly Multicopter Abnormal Data (RflyMAD) is designed to resolve the problem of insufficient data in research fields like FDI and HA. Considering the balance of the amount of data and the real flight data, RflySim platform is used to generate simulation data. RflySim is a model-based platform used to realize rapid development \citep{quan2020multicopter, dai2021rflysim}. A nonlinear multicopter models with high-accuracy and fault injection modules is built in this platform. Together with the PX4 autopilot, concepts of SIL and HIL simulation are applied to collect the SIL and HIL simulation data. Next, with the modified source code of PX4 firmware, the same fault injection modules are used to collect the real flight data. RflyMAD contains different kinds of faults in actuators (motors, propellers), sensors (accelerometer, gyroscope, magnetometer, barometer and GPS) and environment effects. Each fault type has a variety of fault parameters that present the degree of failure. In each case, telemetry log from QGroundControl, ULog from autopilot are both provided, and a CSV containing the concrete flight information is also attached. Besides, the ground truth data with a high frequency from RflySim platform in simulation and ROS bag files in real flight are both recorded. Lastly, some basic usage methods and toolkits are provided to make RflyMAD easier to use. 

The remainder of the paper is organized as follows: \emph{Section \ref{sect_pro}} shows the properties of the RflyMAD dataset. 
%RflyMAD is a large-scale, high quality, and suitable for both simulation and real dataset. 
\emph{Section \ref{sect_construct}} introduces the hardware and software used for collecting the data and how to collect. \emph{Section \ref{sect_val}} verifies the support relationship between the simulation and real flight data. \emph{Section \ref{sect_discuss}} consists of the future work and the problems that exist in the current dataset.

\section{Properties of RflyMAD}\label{sect_pro}
In order to improve both quality and quantity of data, RflyMAD consists of simulation and real flight data. In this section, properties of the dataset will be introduced, including data formats, dataset hierarchy and fault types.

\subsection{Data formats}\label{Pro_format}
Each flight within the dataset contains four types of raw data and relevant processed files, which could be described as the following.

\begin{itemize}[leftmargin=*,parsep=0pt,topsep=0pt,itemsep=0.5pt]
\item \textbf{Flight Information}. It contains the flight command (e.g., take off, land and move to a target position), fault types and fault parameters. The data is provided in format comma-separated values (CSV) called \textit{TestInfo.CSV} in the dataset.
\item \textbf{ULog}. It is used to log uORB topics as messages, including device inputs (e.g., sensors, RC inputs), internal state (e.g., attitude, EKF states), and String messages. The file could be converted into CSV conveniently. This data is provided in original format and processed CSV format.
\item \textbf{Telemetry Log}. TLog is recorded by the ground station, and the main content is the information communicated between a multicopter and its corresponding QGroundControl\footnote[2]{ The QGroundControl website: \href{http://qgroundcontrol.com/}{http://qgroundcontrol.com/}} (QGC). Thus frequency of transmission is decided by the communication quality in real flight or the performance of the simulation computer. This data is provided in format: tlog(original) and CSV (processed).
\item \textbf{Ground Truth Data}. It is generated by RflySim platform during the simulation and recorded at approximately 120Hz. It contains the kinematics information, fault states and motor speeds. This data is abbreviated as ``GTData'' in the following text and is provided in CSV format.
\item \textbf{BAG}. It is generated by the ROS system in each real flight. It contains the position, attitude, and control commands of a multicopter. This data is provided in two ways: the raw data without being processed, and the processed BAG data converted to CSV files.
\end{itemize}

It is worth noting that the GTData only exists in simulation data and the BAG files only in real flight data, so there are still four types of data associated with each flight. All the data mentioned above can be transformed, restored and preprocessed automatically by the toolkits we developed.

\subsection{Scale}\label{Pro_scale}
RflyMAD is a large-scale dataset in research field of UAV abnormal data, for it contains a great amount of simulation data which could be divided into SIL and HIL simulation data in detail. Together with the real flight data, RflyMAD is consisted of three sub-datasets, whereas the fault types and flight statues in each sub-dataset are still similar.

TABLE \ref{tab_fault_type} shows the number of flights in each sub-dataset associated with different fault types. There are 11 fault types in simulation data and 7 in real flight data to cover the common faults that may occur in a multicopter. It is worth noting that data about different combinations of failure units of motor or propeller are collected. Most importantly, faults from the power system, sensor system, multicopter’s frame structure to the environment effects are considered. We believe that the more fault types we have, the more extensive the dataset will become. 

\begin{table}[htbp]
	\small\sf\centering
	\caption{Summary of RflyMAD Fault Types}
	\begin{tabular}{c c c c}
		\toprule
		\textbf{Type of}&\multicolumn{3}{c}{\textbf{Type of Sub-dataset}} \\
		\cline{2-4} \rule{0pt}{10pt}
		\textbf{Faults} & \textbf{\textit{SIL Sim}}& \textbf{\textit{HIL Sim}}& \textbf{\textit{Real Flight}} \\
		\midrule
		Motor (1-4) & 921 & 921 & 231  \\
		Propeller (1-4) & 435 & 435 & $\times$ \\
		Low Voltage & 20 & 20 & $\times$ \\
		Wind Affect & 150 & 150 & $\times$  \\
		Load Lose & 150 & 150 & $\times$  \\
		Sensors' Noise & 50 & 50 & 82 \\
		Accelerometer & 128 & 128 & 20 \\
		Gyroscope & 128 & 128 & 20 \\
		Magnetometer & 128 & 128 & 20 \\
		Barometer & 128 & 128 & 20 \\
		GPS & 128 & 128 & 20 \\
		No Fault & 200 & 200 & 84 \\
		\hline \rule{0pt}{10pt}
		Total & 2566 & 2566 & 497 \\
		\bottomrule
		\multicolumn{4}{l}{Note:$\times$ represents this item does not exist in sub-dataset.  \rule{0pt}{10pt}}\\
		\multicolumn{4}{l}{Motor(1-4) represents the number of failure motors is in}\\
		\multicolumn{4}{l}{range of 1 to 4.}
	\end{tabular}
	\label{tab_fault_type}
\end{table}

In TABLE \ref{tab_fault_type}, there are 5629 flight cases in total, so the total size of RflyMAD is about 114.6 GB. In each flight for one sub-dataset, the fault parameters which represent the failure degree is different even if they have the same fault types. Due to the high cost of real flight with faults injected, the real flight dataset does not contain so much data with high failure degree compared with simulation dataset. 

Considering when the fault occurs, the multicopter may be at different flight statuses and thus have different performance. For the purpose of fully studying the mobility of the multicopter, we design 6 flight statuses when the fault occurs. TABLE \ref{tab_flight_status} shows the flight status in each sub-dataset. The selected flight status refers to the Mission-Task-Elements in the ADS-33 file. The ADS-33 file is a specification document for U.S. military rotorcraft flight quality  \citep{baskett2000aeronautical}. As this document is mainly used for helicopters, some basic tasks are selected and modified to be suitable for multicopters. In each flight status, RflyMAD dataset contains all fault types with different fault parameters in order to satisfy more conditions. 

\begin{table}
	\small\sf\centering
	\caption{Summary of RflyMAD Flight Status}
	\begin{tabular}{c c c c}
		\toprule
		\textbf{Flight}&\multicolumn{3}{c}{\textbf{Type of Sub-dataset}} \\
		\cline{2-4} \rule{0pt}{10pt}
		\textbf{Status} & \textbf{\textit{SIL Sim}}& \textbf{\textit{HIL Sim}}& \textbf{\textit{Real Flight}} \\
		\midrule
		Hover& \checkmark & \checkmark & \checkmark\\
		Waypoints&  \checkmark&  \checkmark & \checkmark\\
		Velocity Control&  \checkmark &  \checkmark & \checkmark \\
		Circling&  \checkmark &  \checkmark & \checkmark \\
		Acceleration&  \checkmark &  \checkmark & \checkmark  \\
		Deceleration&  \checkmark &  \checkmark & $\times$  \\
		\bottomrule
		\multicolumn{4}{l}{Note: \checkmark represents this item exists in sub-dataset and \rule{0pt}{10pt}}\\
		\multicolumn{4}{l}{$\times$ represents not.}\\
	\end{tabular}
	\label{tab_flight_status}
\end{table}

\subsection{Hierarchy}

\begin{figure}[htbp]
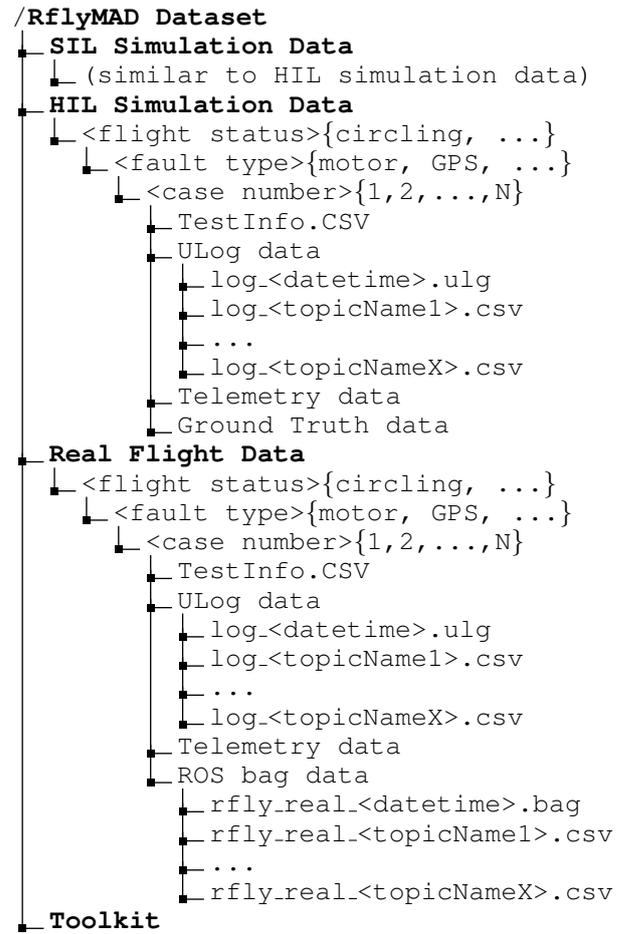

	\setlength{\DTbaselineskip}{11pt}
	\DTsetlength{0.2em}{0.8em}{0.2em}{0.5pt}{3pt}
	\dirtree{%
		.1 /\textbf{RflyMAD Dataset}.
		.2 \textbf{SIL Simulation Data}.
		.3 (similar to HIL simulation data).
		.2 \textbf{HIL Simulation Data}.
		.3 <flight status>\{circling, ...\}.
		.4 <fault type>\{motor, GPS, ...\}.
		.5 <case number>\{1,2,...,N\}.
		.6 TestInfo.CSV.
		.6 ULog data.
		.7 log\_<datetime>.ulg.
		.7 log\_<topicName1>.csv.
		.7 ....
		.7 log\_<topicNameX>.csv.
		.6 Telemetry data.
		.6 Ground Truth data.
		.2 \textbf{Real Flight Data}.
		.3 <flight status>\{circling, ...\}.
		.4 <fault type>\{motor, GPS, ...\}.
		.5 <case number>\{1,2,...,N\}.
		.6 TestInfo.CSV.
		.6 ULog data.
		.7 log\_<datetime>.ulg.
		.7 log\_<topicName1>.csv.
		.7 ....
		.7 log\_<topicNameX>.csv.
		.6 Telemetry data.
		.6 ROS bag data.
		.7 rfly\_real\_<datetime>.bag.
		.7  rfly\_real\_<topicName1>.csv.
		.7 ....
		.7  rfly\_real\_<topicNameX>.csv.
		.2 \textbf{Toolkit}.
	}
	
	\caption{RflyMAD hierarchy.}
	\label{fig_hierarchy}
\end{figure}

The hierarchy of RflyMAD dataset is shown in Figure \ref{fig_hierarchy}. There are five layers from inside to outside of the dataset. The outermost layer comprises three sub-datasets. The second layer is flight status, and there are 6 types for simulation data and 5 types for real flight data. The third layer is fault type, and the fourth layer consists of a series of cases with the same flight status and fault, but with varying fault parameters. The fifth layer consists of four basic files for each flight. The reason why RflyMAD is designed as such a complicated structure is to make data of each flight have a distinct classification. This make it easier for researchers to design model-based or data-driven algorithms. Besides, the fault type, fault parameters and fault injection time could be found in the Flight information, ULog and BAG data in each flight. 

\subsection{Extensibility}\label{pro_SandE}
\begin{comment}
	Different from research fields like computer vision, batteries, and mechanical vibration, there are relatively few publicly datasets related to the FDI problem and  health assessment of aircraft systems. Regarding the lack of fault dataset for fixed wing aircraft, relevant researchers have conducted real flight experiments, forming a dataset called ALFA \citep{keipour2021alfa}. RflyMAD compensates for the shortcomings and deficiencies in this field of multirotor aircraft.
	
	From the perspectives of quality and quantity, on the one hand, with the help of highly reliable simulation platforms and models, as well as some data verification steps, we ensure that the data have high credibility. On the other hand, through specific experimental design and arrangement, we ensured that the dataset had rich diversity, and the total simulation experiment duration was 8000 minutes, with a flight duration of about 1524 minutes in the fault state.
\end{comment}

This paper utilizes the RflySim platform to design and realize simulation experiments, and simultaneously carry out real flight experiments. If the current data contained in RflyMAD cannot meet the application or research needs, users can use the RflySim platform to design and conduct experiments on their own and collect simulation data. Access to the RflySim platform \footnote[3]{RflySim platform website: \href{https://rflysim.com/}{https://rflysim.com/}} is free, and users can learn more about it. With this platfrom, users only need to write the control sequence in advance to arrange the flight status and types of faults to be injected. Once the simulation starts, the data will be collected automatically by programs. And how the simulation and real flight data are collected will be introduced in \emph{Section \ref{sect_construct}}.

\section{Dataset Construction}\label{sect_construct}
RflyMAD dataset is divided into three parts, and each part has its own way to collect. This section will introduce the ways in detail and then show how to construct the dataset.

\subsection{Simulation Data}
\begin{figure}
	\centerline{\includegraphics[width=82mm]{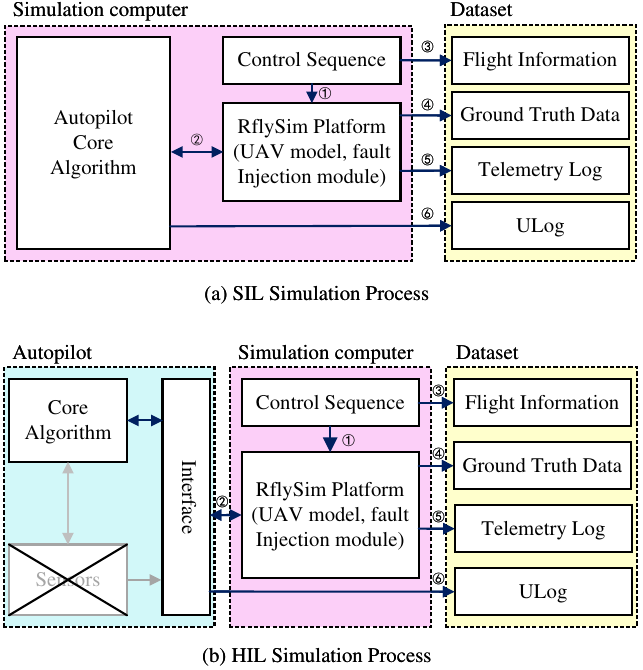}}
	\caption{Methods to collect simulation data.}
	\label{fig_SH}
\end{figure}

Figure \ref{fig_SH}(a) and Figure \ref{fig_SH}(b) show how the simulation data is collected. For SIL simulation data, the hardware equipment is nothing more than a high-performance simulation computer. And only one more Pixhawk autopilot is needed for HIL simulation data \citep{quan2020multicopter,dai2021rflysim,wang2021rflysim}.

In SIL simulation, RflySim is used as our core platform to collect data. In RflySim, we build a nonlinear multicopter dynamic model with high-accuracy compared with the real multicopter. The model has an actuator module, battery module, GPS module, IMU module, and environment module considering air drag and wind effects. Physical parameter of each module is obtained or fitted from real-world data. Besides, we design a fault injection module to change the fault parameters inside the model. In simulation, the fault parameter works by multiplying with throttle signal of motors, sensor signals or adding with them as noises or deviations. The control sequence consists of a series of flight commands and fault information. Flight status of the multicopter, fault types, fault parameters and when the fault will start and come to an end is decided by the control sequence. In the end, the control sequence will be recorded in the dataset as flight information. The core algorithm uses PX4 firmware to control the multicopter model of RflySim in simulation. 

When the simulation begins, the RflySim platform will run the multicopter model and call QGC program, and then execute control sequence sequentially. When the fault is injected, the fault parameters change and the corresponding module loses its function partly or completely fails. Simultaneously, performance of the multicopter becomes abnormal, which indicates the success of fault injection. When the simulation finishes, TLog, ULog, and GTData will be obtained from RflySim platform and restored in the dataset.

In HIL simulation, in addition to the simulation computer, a PX4 autopilot is also needed to work together with the RflySim platform. Compared with the SIL simulation, the HIL simulation keeps core algorithms on autopilot and makes them communicate with the RflySim platform through interface like MAVLink communication protocol. As sensor modules are embedded in the multicopter model of the RflySim platform, the sensor signal from the PX4 autopilot needs to be shielded. So only the core algorithms of PX4 autopilot are used to control the model and generate ULog in autopilot's SD card. Other parts are similar to SIL simulation.

\subsection{Real Flight Data}
\begin{figure}[htbp]
	\centering
	\includegraphics[width=83mm]{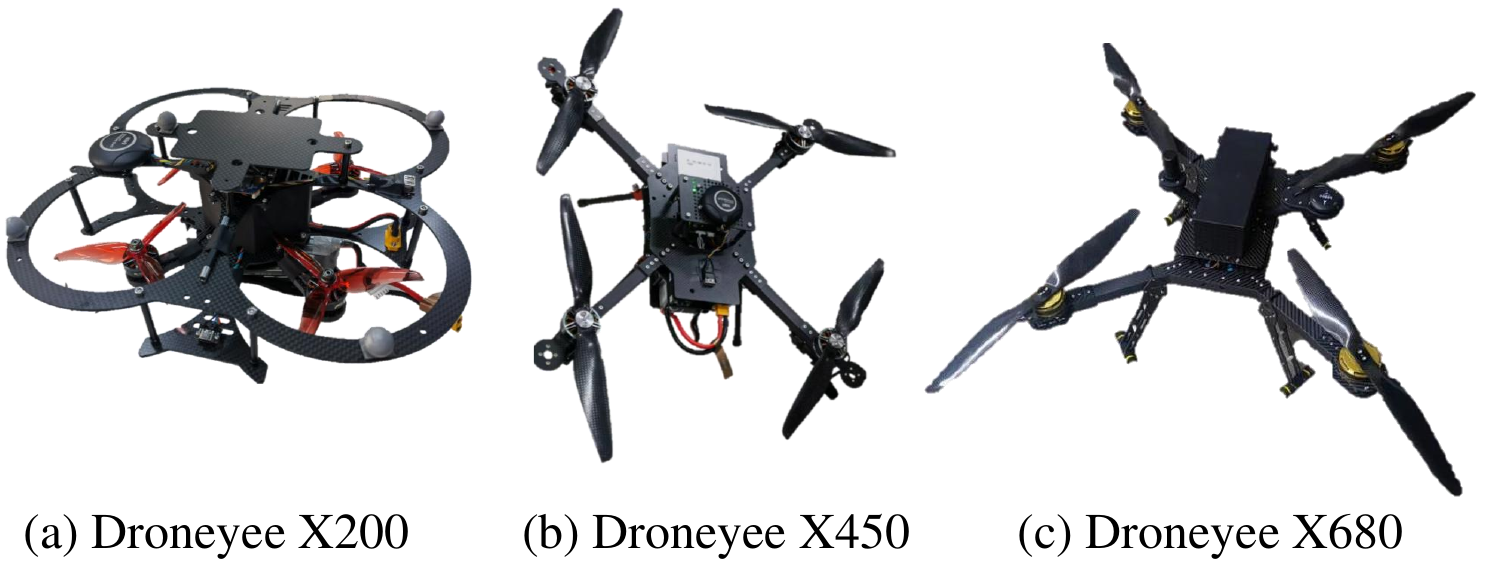}
	\caption{Quadcopters used in real flight.}
	\label{fig_plane}
\end{figure}

Due to the size and weight of the UAV, which may influence the diversity of RflyMAD dataset and the performance of diagnosis methods, there are three kinds of multicopters used to collect real flight data. They are made by Droneyee company with diagonal sizes of 200mm (1.054kg), 450mm (2.084kg) and 680mm (4.068kg). The quadcopters used in real flight can be seen in Figure \ref{fig_plane}. All quadcopters have the same components and sensors, including a PX4 autopilot, a GPS module, an onboard computer, and a radio receiver.

\begin{figure}
	\centering
	\includegraphics[width=80mm]{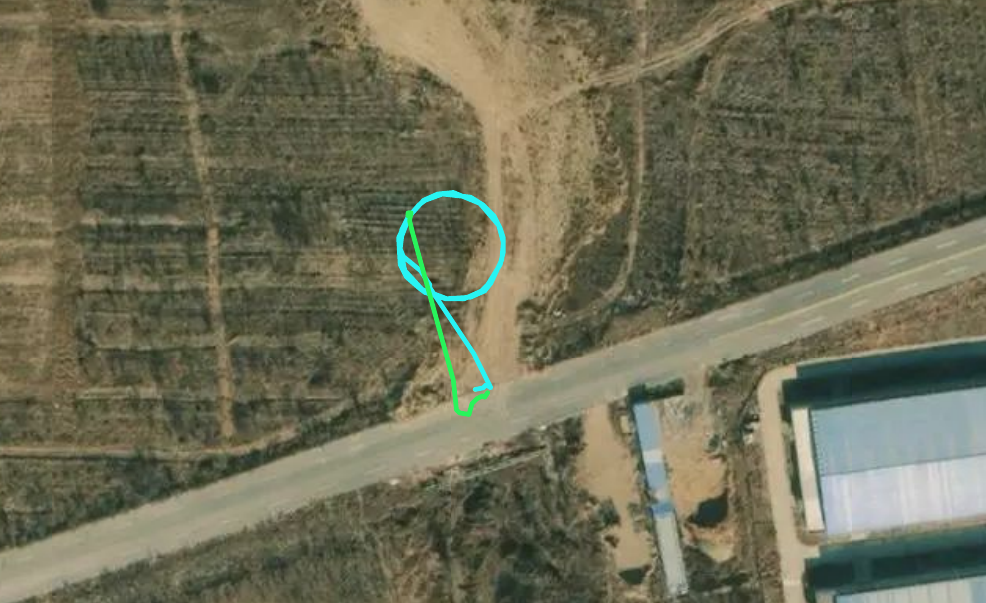}
	\caption{Trajectory of one flight in experiments}
	\label{fig_track}
\end{figure}

\begin{figure}
	\centering
	\includegraphics[width=82mm]{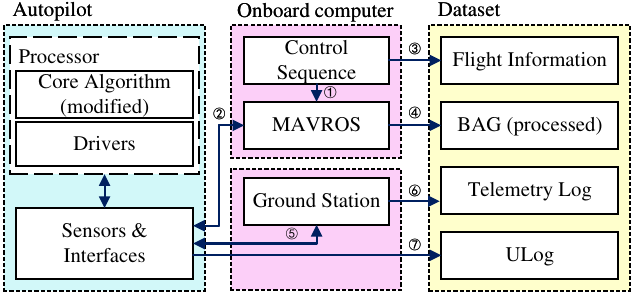}
	\caption{Methods to collect real flight data.}
	\label{fig_R}
\end{figure}

Figure \ref{fig_track} presents one trajectory of the quadcopter in our experiment ground, located in Huailai county, Zhangjiakou city, Hebei province, P. R. China. In our experiments, the quadcopter will follow commands from the control sequence, executing different flight statuses when faults are injected. The detail about flight status and fault types are introduced in \emph{Section \ref{Pro_scale}}.

Figure \ref{fig_R} shows that the construction process of real flight data is quite different from that for simulation data. Firstly, the whole of PX4 autopilot is used without shielding the sensor signals. In real flight, the fault injection module is developed again by modifying the core algorithm of PX4. The firmware version of PX4 used is 1.12.3. The modified methods can be summarized in three steps as follows:

\begin{itemize}[leftmargin=*,parsep=0.5pt,topsep=0.5pt,itemsep=0.5pt]
\item Create a \textit{ctrl.msg} file for uORB topic. This file includes timestamp, control flag, mode flag and control signals. 
\item Subscribe the \textit{ctrl} signals in each module needed to be modified. Copy the fault parameters into each part.
\item Update firmware signals with fault parameters. If the corresponding fault occurs, the fault parameters will be added or multiplied to the original signals.
\end{itemize}

Secondly, the onboard computer uses Robot Operating System (ROS) Noetic Ninjemys with Ubuntu 20.04 system to collect information of flight states, sensor signals, and input control commands through package MAVROS, also called ``MAVLink for ROS''. It could transform ROS topic message into the format of MAVLink and send the message into PX4. Then the message will be converted into the format of uORB and published in PX4. Using this method, the control sequence we have written could be sent into the core algorithm of PX4 and realize fault injection. Similar to simulation data, the control sequence will be collected as ``Flight information'' in the real flight data.

The MAVROS can receive messages from autopilot and generate a data file called ROS bag through ROS in each flight. We automatically collect this data as BAG in the dataset. ULog and TLog data are collected separately from the autopilot and the QGC.

\section{Data Validation}\label{sect_val}
Compared with the real flight data, the acquisition cost of simulation data is relatively low, and the simulation data and real flight data are expected to be consistent, so that the real flight data can be replaced by the simulation data to some extent. Validating this kind of support relationship is a challenge. A transfer learning method \citep{azad2023intelligent,hakim2022systematic} is used to check the transfer ability of two kinds of the data,
%by replacing part of the real flight data with high-quality simulation data to complete the training of the model and examine the average accuracy
so as to judge the equivalency between them. Experiments are implemented on a personal computer with Intel(R) Core(TM) i7-12700H CPU, NVIDIA GeForce RTX 3060 Laptop GPU, 16 GB memory and Windows 11 64-bit system. 

\subsection{Experimental Data and Preprocessing}
For experiments, the ULog in real flight and GTData in simulation are used to diagnosis single-motor fault in the hovering flight status. The fault state refers to the situation that the pulling force of a single motor fails from 100\% to 85\% from a certain moment. The diagonal sizes of the simulated aircraft and the actual aircraft are both 450mm, as shown in Figure \ref{fig_plane}(b).

By using the toolkits we developed, the ULog and GTData are converted into format CSV and ordered by timestamps. Although the different sensors in one data have different starting working time and sampling frequencies, it is necessary to unify the clocks of different sensors and unify the sensor data acquisition time accuracy to 10ms. Then interpolation processing is performed on the sensor sampling data whose accuracy is not satisfied. Finally, each sample contains timestamps and 12 characteristic information such as velocity, angular velocity, acceleration, and Euler angle in three dimentions.

Selecting 600 SIL data samples, 600 HIL data samples, 600 real flight data samples, and the normal and fault data account for 50\% of the samples. In the experiment, 400 pieces are randomly sampled from the SIL simulation data, HIL simulation data and the real flight data respectively described as $D_{SIL}$, $D_{HIL}$ and $D_{R}$. 140 pieces of HIL simulation data and 140 pieces of real flight data are used as two different test set $D_{TestHil}$ and $D_{TestR}$. Accordingly, 60 pieces of real flight data and 60 pieces of HIL simulation data are selected as two different verification sets.

\subsection{Model and Domain Adaptation Algorithms}
\begin{comment}
	In the application of fault diagnosis based on data driven methods, transfer learning can be used to replace part of the real flight data with high-quality simulation data to complete the training of the model, thereby reducing the demand and dependence on real flight data. 
	
	Using the classic transfer learning method, the experimental verification of the support relationship between the simulation data and real flight data was carried out through the flight log file of the UAV, and the support relationship between them was quantitatively analyzed, which provided a basis for the subsequent real flight data collection, simulation data quality evaluation and fault diagnosis algorithm analysis.
\end{comment}
The model used in the experiment is an improved model based on the classic LeNet-5 model  \citep{lecun1998gradient}, and the model parameters are shown in Table \ref{tab_model_param}. The classic TrAdaBoost \citep{dai2007boosting} and AdaBN \citep{li2018adaptive} domain adaptation algorithms in transfer learning are used to verify the support relationship between simulation data and real flight data. By changing the number of samples from $D_{R}$ in the training set and the domain adaptation algorithm, the results of different experiments are compared and analyzed.

\begin{table}
	\small\sf\centering
	\caption{Model Parameters}
	\setlength\tabcolsep{20pt} 
	\begin{tabular}{c c c}
		\toprule
		\textbf{No.} & \textbf{Layers}& \textbf{Output} \\
		\midrule
		1 & Convolution1D & 12$\times$64  \\
		2 & Batch Normalization & 12$\times$64 \\
		3 & Convolution1D & 12$\times$32 \\
		4 & Batch Normalization & 12$\times$32 \\
		5 & Dense & 1$\times$64  \\
		6 & Batch Normalization & 1$\times$64 \\
		7 & Dense & 1$\times$32 \\
		8 & Batch Normalization & 1$\times$32 \\
		9 & Dense & 1$\times$2 \\
		\bottomrule
	\end{tabular}

	\label{tab_model_param}
\end{table}

\begin{figure}
	\centering
		\includegraphics[width=79mm]{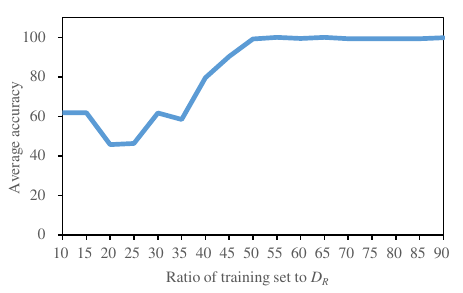}
	\caption{Results of different $D_{R}$ ratios in the training set.}
	\label{fig_results}
\end{figure}

Experiments are implemented using Google's Tensorflow toolbox\footnote[4]{TensorFlow toolbox: \href{https://www.tensorflow.org/}{https://www.tensorflow.org/}}. To minimize the loss function, the Adam optimization algorithm is used to train the model. Epochs are set to 10 in different experiments, and the average prediction accuracy is calculated after multiple training. The experimental results are shown in Table \ref{tab_experi_result} and Figure \ref{fig_results}.

\begin{comment}
	\begin{table}[htbp]
		\small\sf\centering
		\caption{Experimental results}
		\resizebox{\columnwidth}{!}{
			\begin{tabular}{c c c c c}
				\toprule
				\textbf{No.} & \textbf{Training} & \textbf{Test} & \textbf{Adaptive}& \textbf{Average} \\
				\textbf{} & \textbf{set} & \textbf{set} & \textbf{algorithm}& \textbf{accuracy}\\
				\midrule
				1 & $10\%D_{R}\sim90\%D_{R}$ & $D_{TestR}$ & None & 45.7\%$\sim$100\%  \\
				2 & $D_{SIL}$ & $D_{TestR}$ & None & 63\% \\
				3 & $D_{SIL}\cup10\%D_{R}$ & $D_{TestR}$ & None & 96\% \\
				4 & $D_{SIL}\cup10\%D_{R}$ & $D_{TestR}$ & TrAdaBoost & 97\% \\
				5 & $D_{SIL}\cup10\%D_{R}$ & $D_{TestR}$ & AdaBN & 98.2\% \\
				6 & $D_{HIL}$ & $D_{TestR}$ & None & 55.9\% \\
				7 & $D_{HIL}\cup10\%D_{R}$ & $D_{TestR}$ & None & 95.8\% \\
				8 & $D_{HIL}\cup10\%D_{R}$ & $D_{TestR}$ & TrAdaBoost & 99.1\% \\
				9 & $D_{HIL}\cup10\%D_{R}$ & $D_{TestR}$ & AdaBN & 99.1\% \\
				10 & $D_{SIL}$ & $D_{TestHIL}$ & None & 97\% \\
				11 & $D_{SIL}\cup10\%D_{HIL}$ & $D_{TestHIL}$ & None & 97.8\% \\
				12 & $D_{SIL}\cup10\%D_{HIL}$ & $D_{TestHIL}$ & TrAdaBoost & 97.8\% \\
				13 & $D_{SIL}\cup10\%D_{HIL}$ & $D_{TestHIL}$ & AdaBN & 97.8\% \\
				\bottomrule
				\multicolumn{4}{l}{Note: The  average accuracy in experiment 1 is explained in Figure \ref{fig_results}.\rule{0pt}{10pt}}
		\end{tabular}}
		\label{tab_experi_result_discard}
	\end{table}
\end{comment}
\begin{table}
	\small\sf\centering
	\caption{13 Experimental results}
	\setlength\tabcolsep{0.3pt} 
	\resizebox{\columnwidth}{!}{%
		\begin{tabular}{cccccc}
			\toprule
			\large\textbf{No.} & \multicolumn{1}{c}{\large\textbf{\makecell[c]{The support \\ relationship}}} & \large\textbf{Training set} & \multicolumn{1}{c}{\large\textbf{Test set}} & \large\textbf{\makecell[c]{Adaptive \\ algorithm}} & \large\textbf{\makecell[c]{Average \\ accuracy}}  \\
			\midrule
			\large 1 &\multirow{5}{*}{\large \makecell[c]{SIL and \\ real data}} & \large 10\%$D_{R}\sim$90\%$D_{R}$ & \multirow{9}{*}{\large $D_{TestR}$} & \large None & \large 45.7\%$\sim$100\% \\ [2.5pt]
			\large 2 & & \large $D_{SIL}$ & & \large None & \large 63\% \\ [2.5pt]
			\large 3 & & \large $D_{SIL}\cup$10\%$D_{R}$ & & \large None & \large 96\% \\ [2.5pt]
			\large 4 & & \large $D_{SIL}\cup$10\%$D_{R}$ & & \large TrAdaBoost & \large 97\% \\ [2.5pt]
			\large 5 & & \large $D_{SIL}\cup$10\%$D_{R}$ & & \large AdaBN & \large 98.2\% \\ [2.5pt] \cline{2-2}
			\large 6 & \multirow{4}{*}{\large \makecell[c]{HIL and \\ real data}} &\large $D_{HIL}$ & & \large None & \large 55.9\% \\ [2.5pt]
			\large 7 & & \large $D_{HIL}\cup$10\%$D_{R}$ & & \large None & \large 95.8\% \\ [2.5pt]
			\large 8 & & \large $D_{HIL}\cup$10\%$D_{R}$ & & \large TrAdaBoost & \large 99.1\% \\ [2.5pt]
			\large 9 & & \large $D_{HIL}\cup$10\%$D_{R}$ & & \large AdaBN & \large 99.1\% \\ [2.5pt] \cline{4-4}\cline{2-2}
			\large 10 & \multirow{4}{*}{\large \makecell[c]{SIL and \\ HIL data}} & \large $D_{SIL}$ & \multirow{4}{*}{ \large $D_{TestHIL}$} & \large None & \large 97\% \\ [2.5pt]
			\large 11 & & \large $D_{SIL}\cup$10\%$D_{HIL}$ & & \large None & \large 97.8\%  \\ [2.5pt]
			\large 12 & & \large $D_{SIL}\cup$10\%$D_{HIL}$ & & \large TrAdaBoost & \large 97.8\%  \\ [2.5pt]
			\large 13 & & \large $D_{SIL}\cup$10\%$D_{HIL}$ & & \large AdaBN & \large 97.8\%  \\
			\bottomrule
			\multicolumn{6}{l}{\large Note: The  average accuracy in experiment 1 is explained in Figure \ref{fig_results}. \rule{0pt}{12pt}}
		\end{tabular}%
	}
	\label{tab_experi_result}
\end{table}
Through the above experiments, the following conclusions could be drawn.
\begin{itemize}[leftmargin=*,parsep=0.5pt,topsep=0.5pt,itemsep=0.5pt]
\item In \textit{Experiment 1}, with the gradual increase of the data sample size of the $D_{R}$ in the training set, the model prediction accuracy basically exhibits an increasing trend.	
\item In \textit{Experiment 2}, when there is no domain adaptation algorithm and data set $D_{R}$, the model could not apply the feature recognition method learned by $D_{SIL}$, resulting in low prediction accuracy, which also indicated that the data distribution gap between the  SIL simulation data (source domain) and the real flight data (target domain) is large. 	
\item Through \textit{Experiments 2 and 3}, it could be found that after adding target domain data to the training set, the gap between overall data distribution and target domain are reduced, thus improving the prediction accuracy under the same model.
\item In \textit{Experiments 4 and 5}, with the domain adaptation algorithm, the prediction accuracy of the model could reach the same level when 50\% of the training is performed. Therefore, it could be shown that the SIL simulation data quality is high, and the classical domain adaptation method combined with enough simulation data could approximately replace 40\% of the real flight data.
\item Similar results could be seen by replacing SIL simulation data with HIL simulation data in \textit{Experiments 6-9}.
\item In \textit{Experiments 10-13}, it is verified that SIL simulation data could effectively support HIL simulation data in a similar way, and even without HIL simulation data and domain adaptation algorithm, high detection accuracy could still be achieved.
\end{itemize}

\section{Future Work and Discussion} \label{sect_discuss}

\subsection{Discussion}
To the best of our knownledge, there are few high quality datasets with abnormality data for multicopters, or the high quality data are not open-source for some reason. The \href{https://rfly-openha.github.io/documents/4_resources/dataset.html}{RflyMAD dataset} is an open-source dataset that contains both normal and abnormal data with the simulation and real flight for multicopters. 

Although RflyMAD contains many fault types and flight statuses, it is still not enough. First, flight cases for each fault type are not sufficient in real flight sub-dataset. We hope other researchers and enthusiasts would join us and contribute their data with different faults and hardwares by using our collection standards and dataset formats. Besides, the types of data should be diverse. For example, the RflyMAD dataset mainly consists of flight information, ULog, Telemetry log, and processed file from the RflySim platform or ROS system. If we could add more sensors like motor encoders and take videos during the flight with multiple perspectives, the suitably applied fields of RflyMAD would be expanded.

\subsection{Future Work}
We wish the RflyMAD dataset will become an important dataset for multicopter research. So we will continue supporting the development of RflyMAD, providing some basic codes to  facilitate the use of the dataset and to design some model-based and data-driven methods as benchmarks for other researchers to compare and improve their algorithms. As mentioned above, we will add fault types and flight cases in simulation and real flight. The whole dataset and the succeeding data will be publicly available and can be accessed from our website. Besides, efficient toolkits will also be developed to imporove the quality of the dataset.

% It is hard to examine the quality of a set of data, for it is inevitable to use human resources to examine all collected data. In the future, we will develop efficient ways to improve dataset quality with fewer human resources.

\begin{acks}
Thanks to Droneyee company\footnote[5]{Droneyee Company website: \href{http://www.feisilab.com/}{http://www.feisilab.com/}} for supporting equipments and UAVs in real flight experiments. Also, the authors would like to thank Yifan Liu, Haoyu Wei, Xinquan Chen and Ziqiang Wen for their help during months of real flight on hot summer days.
\end{acks}

\bibliography{referfile}

\end{document}